\documentclass[10pt,twocolumn,letterpaper]{article}

\usepackage{cvpr}
\usepackage{times}
\usepackage{epsfig}
\usepackage{graphicx}
\usepackage{amsmath}
\usepackage{amssymb}

\usepackage[pagebackref=true,breaklinks=true,letterpaper=true,colorlinks,bookmarks=false]{hyperref}

\cvprfinalcopy % *** Uncomment this line for the final submission

\def\cvprPaperID{10422} % *** Enter the CVPR Paper ID here

\ifcvprfinal\fi

\usepackage{xspace}
\newcommand{\DCCK}{BSConv\xspace}
\newcommand{\DCCKU}{\DCCK-U\xspace}
\newcommand{\DCCKS}{\DCCK-S\xspace}

\usepackage{cleveref}

\usepackage{xcolor}

\usepackage{amssymb}
\newcommand{\R}{\mathbb{R}}
\newcommand{\tensorIn}{U}
\newcommand{\tensorOut}{V}
\newcommand{\filter}[1]{F^{(#1)}}
\newcommand{\blueprint}[1]{B^{(#1)}}
\newcommand{\blueprintWidthwise}{B'}
\newcommand{\weight}{w}
\newcommand{\weightArray}{\tilde{w}}
\newcommand{\weightA}{w^A}
\newcommand{\weightAArray}{\tilde{w}^A}
\newcommand{\weightB}{w^B}
\newcommand{\weightBArray}{\tilde{w}^B}
\newcommand{\weightMatrix}{W}
\newcommand{\weightMatrixA}{W^A}
\newcommand{\weightMatrixB}{W^B}
\newcommand{\spatialV}{Y}
\newcommand{\spatialW}{X}
\newcommand{\spatialK}{K}

\newcommand{\channelInCount}{M}
\newcommand{\channelInIndex}{m}
\newcommand{\channelOutCount}{N}
\newcommand{\channelOutIndex}{n}

\newcommand{\widthFactor}[1]{(x#1)}

\begin{document}

% space optimization (less space between text and equations)
\setlength{\abovedisplayskip}{5pt}
\setlength{\belowdisplayskip}{5pt}
\setlength{\abovedisplayskip}{5pt}
\setlength{\belowdisplayskip}{5pt}

%%%%%%%%%%%%%%%%%%%%%%%%%%%%%%%%%%%%%%%%
% frontmatter
%%%%%%%%%%%%%%%%%%%%%%%%%%%%%%%%%%%%%%%%

\title{Rethinking Depthwise Separable Convolutions:\\How Intra-Kernel Correlations Lead to Improved MobileNets}

\author{Daniel Haase\thanks{Authors contributed equally}\\
ZEISS Microscopy\\
{\tt\small daniel.haase@zeiss.com}
\and
Manuel Amthor$^*$\\
ZEISS Microscopy\\
{\tt\small manuel.amthor@zeiss.com}
}

\maketitle
\thispagestyle{empty}

%%%%%%%%%%%%%%%%%%%%%%%%%%%%%%%%%%%%%%%%
% abstract
%%%%%%%%%%%%%%%%%%%%%%%%%%%%%%%%%%%%%%%%

\begin{abstract}
We introduce blueprint separable convolutions (\DCCK) as highly efficient building blocks for CNNs.
They are motivated by quantitative analyses of kernel properties from trained models, which show the dominance of correlations along the depth axis.
Based on our findings, we formulate a theoretical foundation from which we derive efficient implementations using only standard layers.
Moreover, our approach provides a thorough theoretical derivation, interpretation, and justification for the application of \textit{depthwise separable convolutions} (DSCs) in general, which have become the basis of many modern network architectures.
Ultimately, we reveal that DSC-based architectures such as MobileNets implicitly rely on cross-kernel correlations, while our \DCCK formulation is based on intra-kernel correlations and thus allows for a more efficient separation of regular convolutions.
Extensive experiments on large-scale and fine-grained classification datasets show that \DCCK{}s clearly and consistently improve MobileNets and other DSC-based architectures without introducing any further complexity.
For fine-grained datasets, we achieve an improvement of up to 13.7 percentage points. 
In addition, if used as drop-in replacement for standard architectures such as ResNets, \DCCK variants also outperform their vanilla counterparts by up to 9.5 percentage points on ImageNet.
Code and models are available under \url{https://github.com/zeiss-microscopy/BSConv}.
\end{abstract}
\vspace{-1.25em}

%%%%%%%%%%%%%%%%%%%%%%%%%%%%%%%%%%%%%%%%
% introduction
%%%%%%%%%%%%%%%%%%%%%%%%%%%%%%%%%%%%%%%%

\section{Introduction}
\textit{Convolutional neural networks (CNNs)} \cite{lecun1989backpropagation,krizhevsky2012imagenet} have become the basis of practically all state-of-the-art approaches for image classification, object detection \cite{girshick2014rich}, semantic segmentation \cite{long2015fully}, and many other applications.
In the past, improvements of CNNs were mainly driven by increasing the model capacity, while at the same time ensuring a proper training behavior \cite{ioffe2015batch,he2016deep,he2016identity}.
Recently, this has led to the development of models with half a billion parameters \cite{huang2018gpipe}.
In practical applications, however, the computational capacity is often limited, especially in mobile and automotive contexts.
This fact has led to another important research direction which focuses on improving model efficiency.
The most prominent approaches are based on \textit{depthwise separable convolutions (DSCs)} \cite{sifre2014rigid,howard2017mobilenets}---building blocks which are motivated by exploiting redundancies of filter weights.

\begin{figure}
	\center
	\includegraphics[width=\columnwidth]{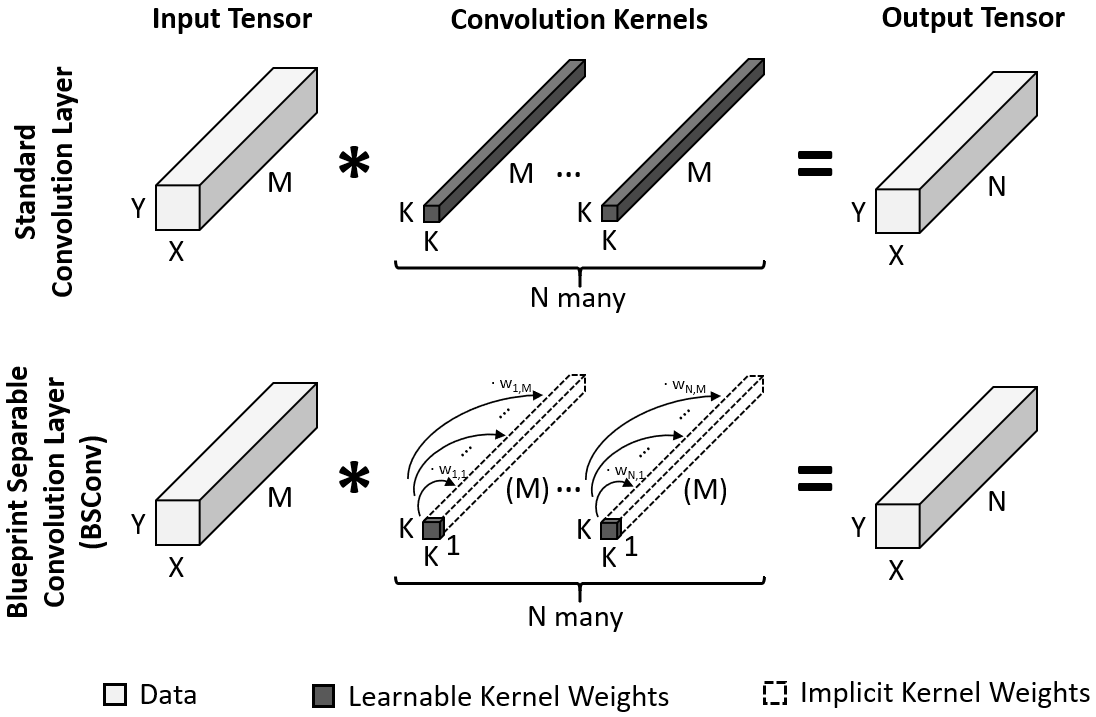}
	\caption{%
		We introduce blueprint separable convolutions (\DCCK) as highly efficient building block for CNNs.
		\DCCK exploits correlations of CNN kernels along their depth axis.
		In consequence, \DCCK represents each filter kernel using one 2d blueprint kernel which is distributed along the depth axis using a weight vector.\vspace{-0.2em}
	}
	\label{fig:convVsDcckSchema}
\end{figure}
\begin{figure*}
	\center
	\includegraphics[width=0.8\textwidth]{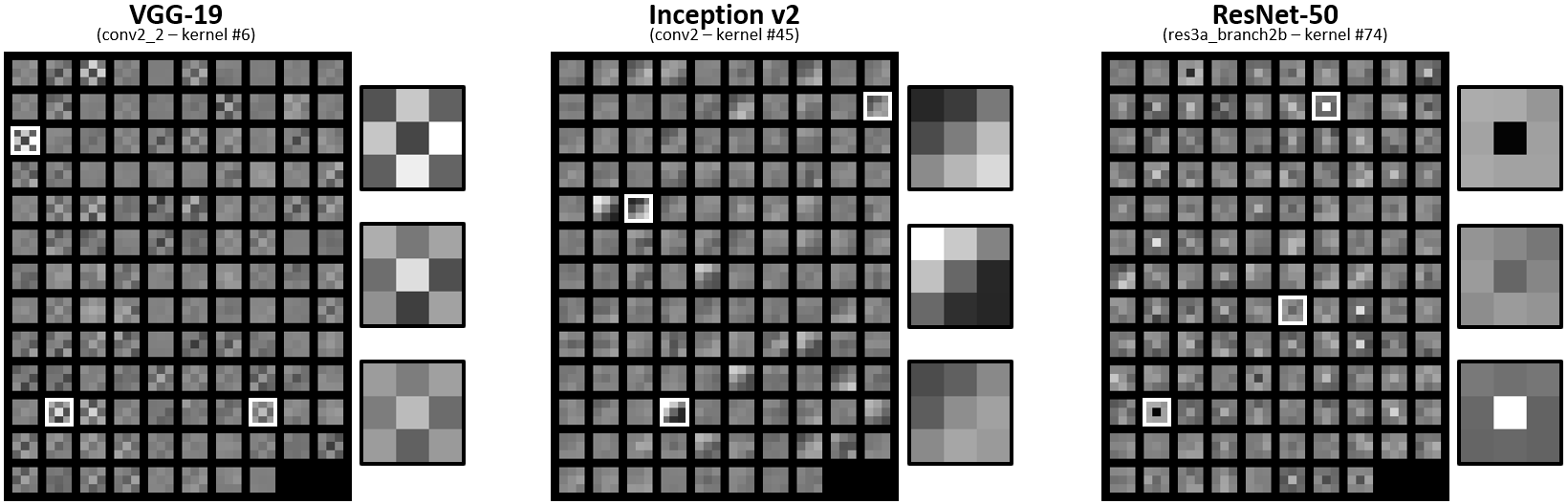}
	\caption{%
		Exemplary filter weights of a vanilla VGG-19, Inception v2, and ResNet-50 CNN trained on ImageNet.
		For each of the three CNN architectures, \textit{one} filter kernel of size $\channelInCount \times \spatialK \times \spatialK = 128 \times 3 \times 3$ is visualized, split into 128 images.
		The weights of each filter kernel are highly correlated along the depth axis.
		Concretely, most slices show the same filter-specific $3 \times 3$ `blueprint', only scaled by different factors (including negative factors which cause `inverted' versions of the blueprint).
		While only three slices are highlighted for each kernel, the correlation is visible for far more slices (52/128 for the ResNet-50 example).
		This observation is the motivation for our proposal of \textit{blueprint separable convolutions (\DCCK)}, which inherently represent a filter kernel of size $\channelInCount \times \spatialK \times \spatialK$ using one blueprint kernel of size $\spatialK \times \spatialK$ and a set of $\channelInCount$ multiplicative factors used to distribute the blueprint across the depth axis.
	}
     \label{fig:W_a}
\end{figure*}

Based on quantitative and qualitative analyses of trained CNNs, in \Cref{sec:dcckOverarching} we propose \textit{blueprint separable convolutions (\DCCK)}, which follow this path of research.
The main idea of \DCCK is to exploit that kernels of CNNs usually show high redundancies along their depth axis (\ie, \textit{intra-kernel correlations}).
Thus, \DCCK represents each filter kernel using one 2d blueprint which is distributed along the depth axis using a weight vector (see \Cref{fig:convVsDcckSchema}).

Although DSCs are also motivated by intra-kernel correlations \cite{sifre2014rigid}, in \Cref{sec:dcck:mobilenets} we show that their derived order of operations contradicts this assumption and is reversed compared to our \DCCK solution.
In fact, the DSC result is equivalent to assuming redundancies between kernels (\ie, \textit{cross-kernel correlations}, see \Cref{fig:depthwiseVsWidthwise}), which was shown to be less effective when separating convolutions \cite{guo2018network}.
In addition, a natural extension of \DCCK leads to an interpretation and justification for DSCs with linear bottlenecks \cite{sandler2018mobilenetv2}, which are extensively used in many recent models \cite{howard2019searching,tan2019mnasnet,tan2019efficientnet}.
Our solution, however, directly implies the use of an additional regularization loss to improve the subspace transform implicitly performed by these bottlenecks.

In \Cref{sec:experiments}, we thoroughly demonstrate that \DCCK consistently outperforms DSC-based architectures such as MobileNets on a wide variety of large-scale and fine-grained datasets at the same parameter and time complexity.
Furthermore, \DCCK can be used as drop-in replacement for standard convolution layers and can be applied to other architectures as well, leading to performance gains while drastically increasing model efficiency.

%%%%%%%%%%%%%%%%%%%%%%%%%%%%%%%%%%%%%%%%
% related work
%%%%%%%%%%%%%%%%%%%%%%%%%%%%%%%%%%%%%%%%

\section{Related Work}
Numerous approaches exist to improve the efficiency of CNNs.
One example is model pruning, where filters and/or connections are removed from CNNs during or after model training \cite{han2015learning,li2016pruning,hassibi1993second,liu2018rethinking}.
Closely related and often combined are quantization \cite{rastegari2016xnor,zhou2017incremental,hubara2017quantized} and compression techniques \cite{gong2014compressing,han2015deep,han2016eie} to accelerate model inference.

% architecture search
Another relevant research area is efficiency-driven CNN architecture search.
It can be performed manually \cite{he2015convolutional} or automatically, \eg via genetic algorithms \cite{xie2017genetic} or via reinforcement learning in the form of neural architecture search \cite{zoph2016neural,cai2018proxylessnas}.
The latter is the basis for the most recent advances in building efficient models such as MnasNet \cite{tan2019mnasnet}, MobileNetV3 \cite{howard2019searching}, and EfficientNet \cite{tan2019efficientnet}.

% building
Building blocks typically consist of activation, normalization, and convolution operations, where the latter bear the greatest potential for efficiency optimizations.
Redundancies in convolution weights of trained CNNs are analyzed in \cite{denil2013predicting,sifre2014rigid,shang2016understanding,guo2018network}.
Approaches to reduce these redundancies are for instance low-rank approximations of the filter kernels \cite{denton2014exploiting,jaderberg2014speeding,jin2014flattened} and the usage of grouped convolutions \cite{zhang2018shufflenet,ma2018shufflenet,xie2017aggregated}.
In \cite{sifre2014rigid}, DSCs are introduced, which form the basis for practically all recent efficient network architectures.
Their direct application can for instance be found in MobileNetV1 \cite{howard2017mobilenets}, factorized CNNs \cite{wang2017factorized}, and Xception \cite{chollet2017xception}.
An extension of DSCs are inverted residual bottlenecks which are introduced in MobileNetV2 \cite{sandler2018mobilenetv2}.
They are used in state-of-the-art efficient architectures, including MnasNet \cite{tan2019mnasnet}, MobileNetV3 
\cite{howard2019searching}, and EfficientNet \cite{tan2019efficientnet}.

%%%%%%%%%%%%%%%%%%%%%%%%%%%%%%%%%%%%%%%%
% DCCK
%%%%%%%%%%%%%%%%%%%%%%%%%%%%%%%%%%%%%%%%

\section{Blueprint Separable Convolutions (\DCCK)}
\label{sec:dcckOverarching}
\begin{figure*}
	\center
	\includegraphics[width=0.315\textwidth]{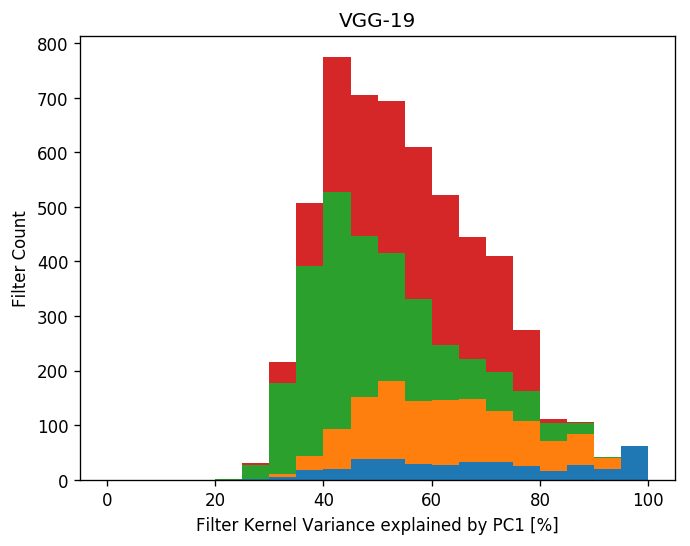}
	\includegraphics[width=0.315\textwidth]{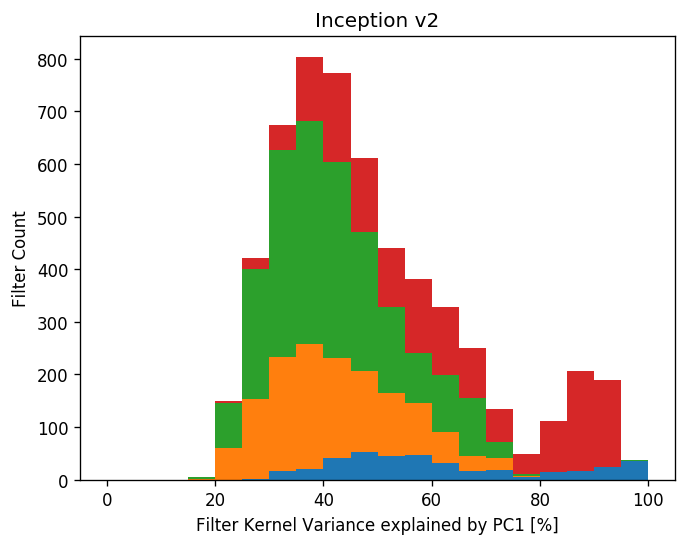}
	\includegraphics[width=0.315\textwidth]{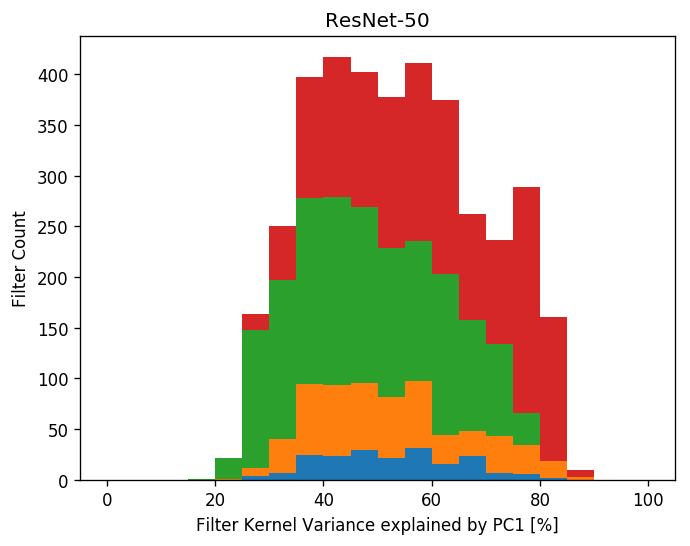}
	\caption{%
		Histogram of the variance along the depth axis of filter kernels which can be explained using only one principal component per filter.
		The filters are grouped by convolution stages (stage 1: blue, stage 2: orange, stage 3: green, stage 4: red).
		These quantitative results show that a large portion of CNN filters can potentially be represented using our \DCCK formulation.
		This figure is best viewed in color.
	}
     \label{fig:correlationsQuantitative}
\end{figure*}

In standard CNNs, each convolution layer transforms an input tensor $\tensorIn$ of size $\channelInCount \times \spatialV \times \spatialW$ into an output tensor $\tensorOut$ of size $\channelOutCount \times \spatialV \times \spatialW$ by applying convolution kernels $\filter{1}, \dots, \filter{\channelOutCount}$, each of size $\channelInCount \times \spatialK \times \spatialK$ such that
\begin{equation}
	\tensorOut_{\channelOutIndex, :, :} = \tensorIn * \filter{\channelOutIndex}
	\label{eq:standardConvolution}
\end{equation}
with $\channelOutIndex \in \left\{1, \dots, \channelOutCount \right\}$ (see \Cref{fig:convVsDcckSchema}, top row).
The entries (or `weights') of these $\channelOutCount$ kernels are optimized during the training stage of CNNs via backpropagation and may take arbitrary values.
However, in the following we show that in practice these weights often converge towards a state in which they show a substantial amount of correlation.
We analyze these correlations qualitatively and quantitatively, and then derive a new, parameter- and time-efficient version of convolution layers for CNNs based on our findings.

\subsection{Intra-Kernel Correlations in Standard CNNs}
\label{subsec:intraKernelCorrelations}
In this paper we focus on intra-kernel correlations and their potential for the design of parameter- and time-efficient CNNs.
We start with a qualitative analysis of intra-kernel correlations by visualizing filters of trained CNNs.
\Cref{fig:W_a} shows exemplary filter kernels for three established CNN architectures trained on the ImageNet dataset---namely VGG-19 \cite{simonyan2014very}, Inception v2 \cite{ioffe2015batch}, and ResNet-50 \cite{he2016deep,he2016identity}.
From these visualizations, it can be seen that intra-kernel correlations exist along the depth axis.
Concretely, it often seems that for a filter $\filter{\channelOutIndex}$, its slices $\filter{\channelOutIndex}_{1, :, :}, \dots, \filter{\channelOutIndex}_{\channelInCount, :, :}$ show the same filter-specific $\spatialK \times \spatialK$ `blueprint', only scaled by different factors (including negative factors which cause `inverted' versions of the blueprint).

While \Cref{fig:W_a} shows only a small subset of filter kernels, the described property of filter slices being based on a `blueprint' are by no means an exception.
According to our observations, it occurs consistently across different CNN architectures, training settings, and datasets.
To systematically quantify to which extent filter kernels show this behavior, we analyze several trained CNNs in the following way:
For each individual filter of the CNN, we
(i) split the $\channelInCount \times \spatialK \times \spatialK$ kernel into $\channelInCount$ samples of size $\spatialK \times \spatialK$,
(ii) perform principal component analysis (PCA) on the set of $\channelInCount$ samples, and
(iii) determine the variance of the filter kernel which is explained by the first principal component (PC1).
Using this approach, we can quantify how well each filter is representable by a $\spatialK \times \spatialK$ filter-specific blueprint (which in this case corresponds to PC1) and $\channelInCount$ factors (which in this case are the `scores' obtained via PCA).
We aggregate these individual values into histograms, which are shown in \Cref{fig:correlationsQuantitative} for the same vanilla CNNs used in \Cref{fig:W_a}.
It can be seen that on average, about 50\% of each filter kernel's variance can be explained using this simple model, suggesting a large potential for efficiency improvements.

\subsection{From Correlations to \DCCK}
\label{subsec:dcck}
The analysis in \Cref{subsec:intraKernelCorrelations} suggests that for trained vanilla CNNs, each $\channelInCount \times \spatialK \times \spatialK$ filter can be approximated using a $\spatialK \times \spatialK$ blueprint and $\channelInCount$ factors which `distribute' the blueprint along the depth dimension.
Even though it is by no means enforced during training, this approximation explains a large portion of the observed variance.
This finding is the motivation for the introduction of \textit{blueprint separable convolutions (\DCCK)}.
They are defined in such a way that above approximation turns into an integral property of the filters of CNNs.
Concretely, we define each filter kernel $\filter{\channelOutIndex}$ to be represented using a blueprint $\blueprint{\channelOutIndex}$ and the weights $\weight_{\channelOutIndex, 1}, \dots, \weight_{\channelOutIndex, \channelInCount} \in \R$ via
\begin{equation}
\filter{\channelOutIndex}_{\channelInIndex, :, :} = \weight_{\channelOutIndex, \channelInIndex} \cdot \blueprint{\channelOutIndex},
\label{eq:dcck}
\end{equation}
with $\channelInIndex \in \left\{1, \dots, \channelInCount \right\}$ and $\channelOutIndex \in \left\{1, \dots, \channelOutCount \right\}$ (see \Cref{fig:convVsDcckSchema}, bottom row).
While this definition poses a hard constraint on the filter kernels, in \Cref{sec:experiments} we experimentally show that CNNs trained with \DCCK can reach the same or even better quality in comparison to their vanilla counterparts.
However, in contrast to standard convolution layers which have $\channelInCount \cdot \channelOutCount \cdot \spatialK^2$ free parameters (see \Cref{fig:convVsDcckSchema}), the \DCCK variant only has $\channelOutCount \cdot \spatialK^2$ parameters for the blueprints, and $\channelInCount \cdot \channelOutCount$ parameters for the weights.
As is discussed below, the latter can be reduced even further.

\subsection{Variants and Implementations}
\label{subsec:dcckVariants}
A \DCCK module consists of $\channelOutCount$ filters, each having one blueprint and $\channelInCount$ weights.
All $\channelInCount \cdot \channelOutCount$ weights can be combined into the matrix $\weightMatrix = (\weight_{\channelOutIndex, \channelInIndex})$.
Depending on how $\weightMatrix$ is learned in the training step, different \DCCK variants can be derived.
In the following, two variants are described.

\subsubsection{Unconstrained \DCCK (\DCCKU)}
\label{subsubsec:dcckU}
\begin{figure}
	\center
	\includegraphics[width=0.975\columnwidth]{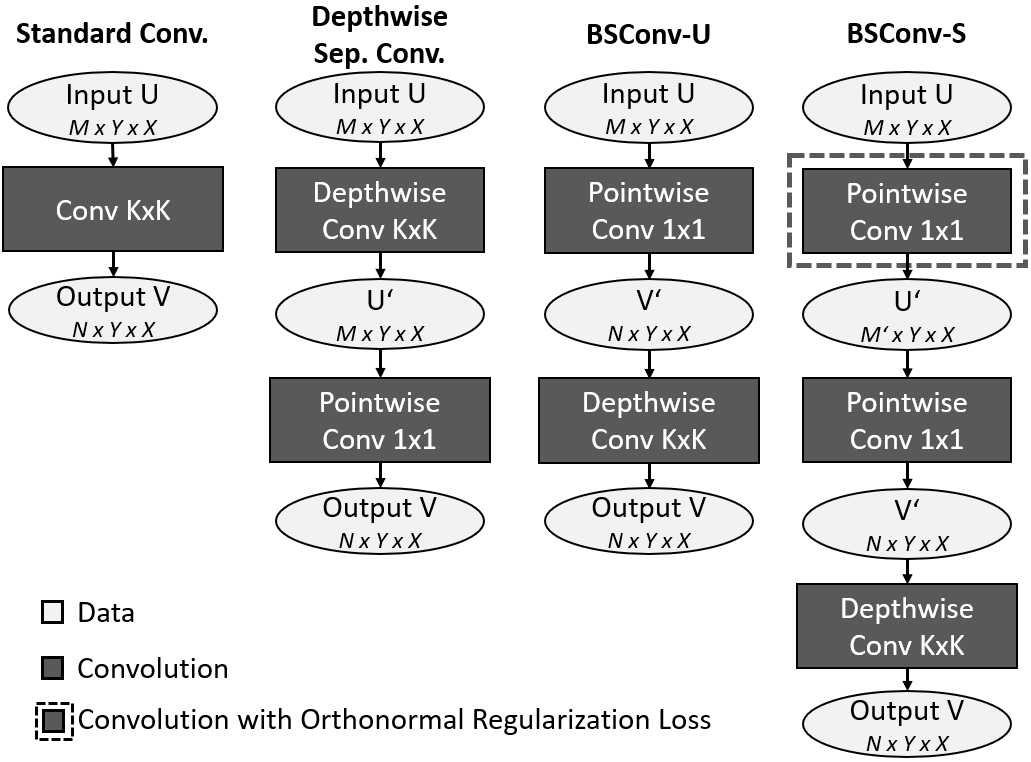}
	\caption{%
		Computational graphs for the efficient implementation of the different \DCCK variants (see \Cref{subsubsec:dcckU,subsubsec:dcckS}).
	}
     \label{fig:cnnImplementationsDcck}
\end{figure}
In the most general case, the weights $\weightMatrix$ can vary without any constraints and are learned directly via backpropagation, just like the entries of the blueprint filter kernels.

As shown in \Cref{fig:convVsDcckSchema}, a naive implementation can be achieved by constructing a full kernel from each blueprint and performing a regular convolution afterwards.
However, to derive a much more efficient implementation for CNNs, we rewrite \Cref{eq:standardConvolution} in the following way:
Firstly, because the input data tensor $\tensorIn$ and the filter kernels $\filter{1}, \dots, \filter{\channelOutCount}$ have the same size $\channelInCount$ along their depth dimension, we can split each 3d convolution into a sum of $\channelInCount$ 2d convolutions, yielding
\begin{equation}
	\tensorOut_{\channelOutIndex, :, :} = \tensorIn * \filter{\channelOutIndex} = \sum_{\channelInIndex = 1}^{\channelInCount} \left( \tensorIn_{\channelInIndex, :, :} * \filter{\channelOutIndex}_{\channelInIndex, :, :} \right) .
\end{equation}
Secondly, we can replace each filter with its \DCCK representation as given in \Cref{eq:dcck}, and obtain
\begin{equation}
	\tensorOut_{\channelOutIndex, :, :} = \sum_{\channelInIndex = 1}^{\channelInCount} \left( \tensorIn_{\channelInIndex, :, :} * \left(  \weight_{\channelOutIndex, \channelInIndex} \cdot \blueprint{\channelOutIndex} \right) \right).
\end{equation}
Because (i) each filter blueprint $\blueprint{\channelOutIndex}$ is independent of the input channel $\channelInIndex$, and (ii) $\weight_{\channelOutIndex, \channelInIndex}$ is a scalar, we can rearrange above equation into
\begin{equation}
	\tensorOut_{\channelOutIndex, :, :} = \underbrace{ \Big( \sum_{\channelInIndex = 1}^{\channelInCount} \tensorIn_{\channelInIndex, :, :} \cdot \weight_{\channelOutIndex, \channelInIndex} \Big) }_{\tensorOut_{\channelOutIndex, :, :}'} * \blueprint{\channelOutIndex}.
	\label{eq:dcckV1Intermediate}
\end{equation}
If we further rearrange $\weight_{\channelOutIndex, 1}, \dots, \weight_{\channelOutIndex, \channelInCount}$ into a $\channelInCount \times 1 \times 1$ array $\weightArray_{\channelOutIndex}$, the sum can be replaced by a convolution, giving
\begin{align}
	\tensorOut_{\channelOutIndex, :, :}' & = \tensorIn * \weightArray_{\channelOutIndex} \label{eq:dcckV1Pointwise} \\
	\tensorOut_{\channelOutIndex, :, :}  & = \tensorOut_{\channelOutIndex, :, :}' * \blueprint{\channelOutIndex}. \label{eq:dcckV1Depthwise}
\end{align}
For a concrete implementation, these \Cref{eq:dcckV1Pointwise,eq:dcckV1Depthwise} can directly be translated into two tensor operations:
(i) a $1 \times 1$ pointwise convolution with the kernels $\weightArray_{1}, \dots, \weightArray_{\channelOutCount}$, which is performed on the input data tensor $\tensorIn$, and
(ii) a $\spatialK \times \spatialK$ depthwise convolution \cite{howard2017mobilenets,ioffe2015batch} with kernels $\blueprint{1}, \dots, \blueprint{\channelOutCount}$, which is applied to the result of the first step.
A flowchart visualization of these steps is given in \Cref{fig:cnnImplementationsDcck}.

\subsubsection{Subspace \DCCK (\DCCKS)}
\label{subsubsec:dcckS}
For CNNs trained with \DCCKU convolution layers, the free parameters to be estimated are the $\channelOutCount$ blueprint kernel matrices and the weight matrix $\weightMatrix = (\weight_{\channelOutIndex, \channelInIndex})$ of size $\channelOutCount \times \channelInCount$.
When analyzing the structure of the weight matrices $\weightMatrix$ of such CNNs, we observe that the rows of $\weightMatrix$ are often highly correlated.
This fact indicates the potential for a further regularization and parameter reduction.

Concretely, we perform a low-rank approximation of $\weightMatrix$ by factorizing it into the $\channelOutCount \times \channelInCount'$ matrix $\weightMatrixA = (\weightA_{\channelOutIndex, \channelInIndex'})$ and the $\channelInCount' \times \channelInCount$ matrix $\weightMatrixB = (\weightB_{\channelInIndex', \channelInIndex})$ as
\begin{equation}
	\weightMatrix = \weightMatrixA \cdot \weightMatrixB,
\end{equation}
with $\channelInCount' = \lceil p \cdot \channelInCount \rceil$, while $p \in (0.0, 1.0)$ specifies the size of the subspace $\channelInCount'$ in relation to the size $\channelInCount$ of the original space.
The matrix $\weightMatrixB$ can be thought of as set of $\channelInCount'$ basis vectors, while $\weightMatrixA$ is the subspace version of $\weightMatrix$.
Instead of $\channelOutCount \cdot \channelInCount$ weights as for the case of \DCCKU, this method reduces the parameter count to $\channelOutCount \cdot \channelInCount' + \channelInCount' \cdot \channelInCount$, as only $\weightMatrixA$ and $\weightMatrixB$ have to be learned via backpropagation.

To minimize redundancies in the low-rank subspace, we want the basis defined by $\weightMatrixB$ to be orthonormal.
This can be achieved via the regularization loss
\begin{equation}
	L_{\text{ortho}} = \big\lVert \weightMatrixB {\weightMatrixB}^{\text{T}} - I \big\rVert_{\text{F}},
\end{equation}
where $I$ is the identity matrix and $\lVert \cdot \rVert_{\text{F}}$ the Frobenius norm of a matrix.
The regularization loss is added to the classification loss $L_{\text{class}}$ with the weighting factor $\alpha$, resulting in the joint loss $L = L_{\text{class}} + \alpha L_{\text{ortho}}$.

To derive an efficient implementation of \DCCKS, we can substitute $\weight_{\channelOutIndex, \channelInIndex} = \sum_{\channelInIndex' = 1}^{\channelInCount'} \weightA_{\channelOutIndex, \channelInIndex'} \cdot \weightB_{\channelInIndex', \channelInIndex} $ in \Cref{eq:dcckV1Intermediate}, yielding
\begin{equation}
	\tensorOut_{\channelOutIndex, :, :} = \Bigg( \sum_{\channelInIndex = 1}^{\channelInCount} \tensorIn_{\channelInIndex, :, :} \cdot \Big( \sum_{\channelInIndex' = 1}^{\channelInCount'} \weightA_{\channelOutIndex, \channelInIndex'} \cdot \weightB_{\channelInIndex', \channelInIndex} \Big) \Bigg) * \blueprint{\channelOutIndex}.
\end{equation}
Using the same arguments as in \Cref{subsubsec:dcckU}, we can rearrange this equation into
\begin{equation}
	\tensorOut_{\channelOutIndex, :, :} = \overbrace{ \Bigg( \sum_{\channelInIndex' = 1}^{\channelInCount'} \underbrace{ \Big( \sum_{\channelInIndex = 1}^{\channelInCount} \tensorIn_{\channelInIndex, :, :} \cdot \weightB_{\channelInIndex', \channelInIndex} \Big) }_{\tensorIn'_{\channelInIndex', :, :}} \cdot \weightA_{\channelOutIndex, \channelInIndex'} \Bigg) }^{\tensorOut'_{\channelOutIndex, :, :}} * \blueprint{\channelOutIndex}.
\end{equation}
By rearranging the weights $\weightB_{\channelInIndex', 1}, \dots, \weightB_{\channelInIndex', \channelInCount}$ into the $\channelInCount \times 1 \times 1$ array $\weightBArray_{\channelInIndex'}$ and the weights $\weightA_{\channelOutIndex, 1}, \dots, \weightA_{\channelOutIndex, \channelInCount'}$ into the $\channelInCount' \times 1 \times 1$ array $\weightAArray_{\channelOutIndex}$, the sums can be replaced by convolutions in the same way as in \Cref{subsubsec:dcckU}, and we obtain
\begin{align}
	\tensorIn'_{\channelInIndex', :, :}  & = \tensorIn * \weightBArray_{\channelInIndex'} \label{eq:dcckV2Pointwise1} \\
	\tensorOut'_{\channelOutIndex, :, :} & = \tensorIn' * \weightAArray_{\channelOutIndex} \label{eq:dcckV2Pointwise2} \\
	\tensorOut_{\channelOutIndex, :, :}  & = \tensorOut'_{\channelOutIndex, :, :} * \blueprint{\channelOutIndex}. \label{eq:dcckV2Depthwise}
\end{align}
Again, \Cref{eq:dcckV2Pointwise1,eq:dcckV2Pointwise2,eq:dcckV2Depthwise} directly translate into tensor operations, thus a concrete implementation of \DCCKS is a three-step process:
(i) the input tensor is projected into a $\channelInCount'$-dimensional subspace via a $1 \times 1$ pointwise convolution with kernels $\weightBArray_1, \dots, \weightBArray_{\channelInCount'}$,
(ii) another $1 \times 1$ pointwise convolution with kernels $\weightAArray_1, \dots, \weightAArray_{\channelOutCount}$ is applied to the result of the first step, and
(iii) a $\spatialK \times \spatialK$ depthwise convolution with kernels $\blueprint{1}, \dots, \blueprint{\channelOutCount}$ is applied to the result of step two (see \Cref{fig:cnnImplementationsDcck} for a visualization).

%\subsubsection{Constant \DCCK (\DCCKC, \DCCKCOne)}
%\label{subsubsec:dcckC}
%The third \DCCK variant we present is the most extreme in terms of parameter reduction and time efficiency.
%However, while \DCCKU and \DCCKS usually match or even improve the performance of comparable vanilla CNNs, for \DCCKC this is generally not the case.
%Therefore, it is mostly suited for scenarios having very low requirements regarding the capacity of CNNs, \eg for very easy datasets.
%
%For \DCCKC, the weight matrix $\weightMatrix$ is assumed to be constant, which completely removes it from the backpropagation process and leaves only the $\channelOutCount \cdot \spatialK^2$ parameters of the blueprints to be learned.
%The concrete implementation for this case is identical to \DCCKU, \ie, performed via a $1 \times 1$ pointwise convolution, followed by a $\spatialK \times \spatialK$ depthwise convolution.
%An even more extreme case, \DCCKCOne, conditions all entries of $\weightMatrix$ to be one.
%As can be seen from \Cref{eq:dcckV1Intermediate}, this reduces the implementation to a summation of the input tensor $\tensorIn$ along the depth axis to form a tensor of size $1 \times \spatialV \times \spatialW$, followed by $\channelOutCount$ 2d convolutions with the kernels $\blueprint{1}, \dots, \blueprint{\channelOutCount}$ (see \Cref{fig:cnnImplementationsDcck}).
%

%%%%%%%%%%%%%%%%%%%%%%%%%%%%%%%%%%%%%%%%
% rethinking
%%%%%%%%%%%%%%%%%%%%%%%%%%%%%%%%%%%%%%%%

\section{Rethinking Depthwise Separable Convolutions}
\label{sec:dcck:mobilenets}
In the following, we show how the derived variants of \DCCK are related to both most important building blocks for mobile models, \ie depthwise separable convolutions and linear inverted residual bottlenecks.
Moreover, we demonstrate how current model architectures can easily be equipped with our improved building blocks.
As we will see in our experiments in \Cref{sec:experiments}, \DCCK variants substantially outperform their vanilla counterparts.

\subsection{\DCCKU is a Reversed Depthwise Separable Convolution}
\label{sec:dcck:mobilenets:dcck-u}

\begin{figure}
	\center
	\includegraphics[width=\columnwidth]{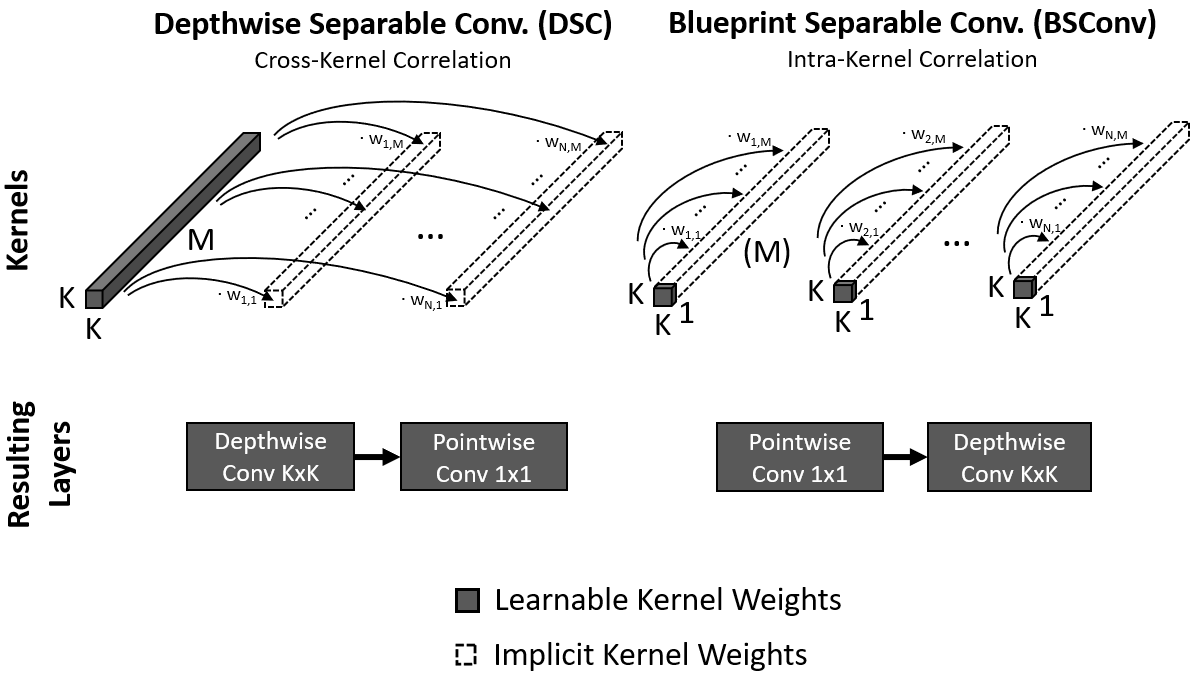}
	\caption{%
		Interpretations for DSC and \DCCK.
		DSCs implicitly assume one 3d blueprint which is used for all kernels, while \DCCK relies on individual 2d blueprints for each kernel.
	}
	\label{fig:depthwiseVsWidthwise}
\end{figure}

Although the derivation of DSCs in \cite{sifre2014rigid} is based on the same observation of kernel correlations along the depth axis, they obtain a reversed order of depthwise and pointwise convolution layers compared to \DCCK (see \Cref{fig:cnnImplementationsDcck}).
This is because DSC in fact enforces cross-kernel correlations instead of intra-kernel correlations.
Using our formulation in \Cref{subsec:dcckVariants}, this can be verified by setting \Cref{eq:dcck} to $\filter{\channelOutIndex}_{\channelInIndex, :, :} = \weight_{\channelOutIndex, \channelInIndex} \cdot \blueprintWidthwise_{\channelInIndex, :, :}$.
This case corresponds to having a single 3d $\channelInCount \times \spatialK \times \spatialK$ blueprint kernel $\blueprintWidthwise$, which is replicated along the width axis, \ie across kernels (see \Cref{fig:depthwiseVsWidthwise}).
While both cross-kernel and intra-kernel correlations are valid assumptions, in \cite{guo2018network} it is shown that the latter dominate and thus have a larger potential for an efficient separation.
This becomes even more obvious given that natural images are inherently correlated along the depth axis, which propagates through all layers.

The MobileNetV1 architecture can be translated to a \DCCKU model by simply substituting all DSCs by \DCCKU building blocks, which effectively means switching the order depthwise and pointwise convolutions.
However, the inversion of the layer order should have no substantial effect on the middle flow of the network, since we already have alternating point- and depthwise convolutions.
The main difference comes from the entry flow: with our approach, feature maps from the first regular convolution can be fully utilized by the depthwise convolution via the preceding pointwise distribution.
In contrast, each kernel of the first depthwise convolution of the original MobileNetV1 model can only benefit from a single feature map, leading to limited expressiveness.
Following our derivation, for the \DCCKU version of MobileNetV1, no activation nor normalization is applied after the pointwise convolutions, since it is essential to allow the weights $\weight_{\channelOutIndex, \channelInIndex}$ to be negative.

\subsection{\DCCKS is a Shifted Linear Bottleneck with Orthonormal Regularization}
\label{sec:dcck:mobilenets:dcck-s}
Linear bottlenecks with inverted residual skips were first introduced in \cite{sandler2018mobilenetv2} as a highly efficient building block providing impressive expressiveness at a minimum amount of required operations.
It follows the idea of regularizing bottlenecks in very large ResNets and became the de facto standard in most of the current state-of-the-art mobile architectures such as MobileNetV2 and MobileNetV3.
A single block consists of a cascade of a pointwise, a depthwise, and a pointwise convolution where the bottleneck is placed between blocks.
Considering \DCCKS, we can see the close relationship when shifting our cascade of two pointwise and one depthwise convolution within a block to obtain a pointwise, a depthwise, and a pointwise convolution.
Thus, MobileNetV2 and V3 can be reinterpreted as a \DCCK model with subspace transforms.
As found by \cite{sandler2018mobilenetv2}, the shift of residual skips into the bottleneck provides superior model performance and we keep that idea also for our implementation of MobileNets equipped with \mbox{\DCCKS}.
Following our derivation in \Cref{subsubsec:dcckS}, the linear bottleneck, \ie\ without the use of an activation function, comes quite natural since negative components are equally essential as positive ones for the subspace transform.
Note that this implies that the first bottleneck block does not apply a subspace transform while the last feature map before the final classification layer is in fact compressed.
Most importantly, \DCCKS models greatly benefit from our theoretical findings concerning the application of the orthonormal regularization during training.
Note that the conversion to a \DCCKS model as described above is also suited for other architectures which use linear bottleneck blocks, such as EfficientNet \cite{tan2019efficientnet} and MnasNet \cite{tan2019mnasnet}.

%%%%%%%%%%%%%%%%%%%%%%%%%%%%%%%%%%%%%%%%
% experiments
%%%%%%%%%%%%%%%%%%%%%%%%%%%%%%%%%%%%%%%%

\section{Experiments}
\label{sec:experiments}
We evaluate our approach of blueprint separable convolutions based on a variety of commonly used benchmark datasets.
We provide a comprehensive analysis of the \mbox{MobileNet} family and their modified counterparts according to our findings in \Cref{sec:dcck:mobilenets}.
Furthermore, we demonstrate how our approach can be used as a drop-in substitution for regular convolution layers in standard models like ResNets to drastically reduce the number of model parameters and operations, while keeping or even gaining accuracy.

To allow for a fair comparison, we train all models---including the baseline networks---with exactly the same training procedure.

\subsection{CIFAR10 and CIFAR100}
The CIFAR10/100 datasets \cite{krizhevsky2009learning} consist of 50k training and 10k test images of size $32\text{\,px} \times 32\text{\,px}$ and comprise 10 and 100 classes, respectively.
As suggested in \cite{he2016deep,zagoruyko2016wide}, we train for 200 epochs for both datasets.
We use SGD with momentum set to $0.9$ and a weight decay of $10^{-4}$.
The initial learning rate is set to $0.1$ and decayed by a factor of $0.1$ at epochs $100$, $150$, and $180$.
Images are augmented by random horizontal flips and random shifts by up to $4$ px to prevent models from overfitting \cite{he2016deep, zagoruyko2016wide}.

\textbf{MobileNets.}
As first experiment, we evaluate our improvements of MobileNetV1--V3 \cite{howard2017mobilenets,sandler2018mobilenetv2,howard2019searching} as derived in \Cref{sec:dcck:mobilenets}.
To make MobileNets applicable to CIFAR, we removed the first and second pooling operation to obtain a final feature map of size $4 \times 4$.
With this modification, we achieve state-of-the-art performance for the baseline models (see \Cref{tab:CIFAR-Mobilenet}).

As described in \Cref{sec:dcck:mobilenets}, we use \DCCKU for \mbox{MobileNetV1} and \DCCKS for MobileNetV2/V3.
For the \DCCKS models, we use a subspace compression ratio of $p = \frac{1}{6}$ to have exactly the same parameter count as the vanilla model.
The weighting coefficient $\alpha$ for the orthonormal regularization loss was set to $0.01$.

\begin{table*}[tb]
	\begin{center}
		\begin{tabular}{|l|c|c||c|c||c|c||c|c||c|c|}
			\hline
			Network & \multicolumn{2}{|c||}{CIFAR10} & \multicolumn{2}{|c||}{CIFAR100} & \multicolumn{2}{|c||}{Stanford Dogs} & \multicolumn{2}{|c||}{Stanford Cars}  & \multicolumn{2}{|c|}{Oxford Flowers} \\
			 &  orig & ours & orig & ours & orig & ours & orig & ours & orig & ours \\
			\hline
			\hline
			MobileNetV1 \widthFactor{0.25} 			& 90.4 & \bf{91.6} & 67.5 & \bf{69.8} & 42.8 & \bf{49.1} & 64.6 & \bf{74.0} & 59.2 & \bf{68.0} \\
			MobileNetV1 \widthFactor{0.5} 			& 91.8 & \bf{93.3} & 70.8 & \bf{73.5} & 49.3 & \bf{55.2} & 70.6 & \bf{78.8} & 63.1 & \bf{71.5} \\
			MobileNetV1 \widthFactor{0.75}			& 92.7 & \bf{94.3} & 72.2 & \bf{74.5} & 51.4 & \bf{57.9} & 72.9 & \bf{80.0} & 63.1 & \bf{70.8} \\
			MobileNetV1 \widthFactor{1.0} 			& 93.4 & \bf{94.3} & 73.4 & \bf{75.7} & 51.6 & \bf{59.1} & 74.4 & \bf{79.9} & 60.2 & \bf{67.3} \\
			\hline
			MobileNetV2 \widthFactor{0.25} 			& 89.6 & \bf{90.1} & 65.6 & \bf{68.9} & 42.0 & \bf{46.8} & 65.2 & \bf{69.9} & 44.9 & \bf{51.9} \\
			MobileNetV2 \widthFactor{0.5} 			& 92.0 & \bf{93.2} & 72.5 & \bf{73.2} & 50.8 & \bf{54.8} & 70.4 & \bf{78.0} & 57.6 & \bf{60.6} \\
			MobileNetV2 \widthFactor{0.75}			& 93.1 & \bf{93.9} & 73.2 & \bf{75.0} & 53.5 & \bf{59.0} & 73.4 & \bf{82.0} & 55.7 & \bf{71.5} \\
			MobileNetV2 \widthFactor{1.0} 			& 93.6 & \bf{94.2} & 74.9 & \bf{75.8} & 56.0 & \bf{60.1} & 76.7 & \bf{83.8} & 61.3 & \bf{67.0} \\
			\hline
			MobileNetV3-small \widthFactor{0.35} 	& 90.3 & \bf{90.6} & 66.5 & \bf{67.2} & 42.8 & \bf{44.2} & 63.4 & \bf{70.4} & 56.9 & \bf{66.5} \\
			MobileNetV3-small \widthFactor{0.5} 		& 91.5 & \bf{91.7} & 69.4 & \bf{69.6} & 45.3 & \bf{47.4} & 68.1 & \bf{74.4} & 64.0 & \bf{71.7} \\
			MobileNetV3-small \widthFactor{0.75}		& 92.0 & \bf{92.5} & 70.4 & \bf{72.0} & 46.7 & \bf{49.5} & 72.1 & \bf{77.2} & 66.3 & \bf{74.3} \\
			MobileNetV3-small \widthFactor{1.0} 		& 92.2 & \bf{92.7} & 72.2 & \bf{73.7} & 49.4 & \bf{52.1} & 72.5 & \bf{77.0} & 68.4 & \bf{75.6} \\
			\hline
			MobileNetV3-large \widthFactor{0.35} 	& 92.8 & \bf{93.0} & 71.5 & \bf{73.7} & 48.5 & \bf{56.0} & 69.5 & \bf{77.5} & 55.7 & \bf{69.4} \\
			MobileNetV3-large \widthFactor{0.5} 		& 93.0 & \bf{93.9} & 72.9 & \bf{75.3} & 51.2 & \bf{57.9} & 73.6 & \bf{80.4} & 65.7 & \bf{66.8} \\
			MobileNetV3-large \widthFactor{0.75}		& 93.7 & \bf{94.4} & 73.9 & \bf{77.0} & 51.8 & \bf{60.0} & 74.9 & \bf{80.9} & 63.1 & \bf{75.1} \\
			MobileNetV3-large \widthFactor{1.0} 		& 93.7 & \bf{94.6} & 75.2 & \bf{77.7} & 54.9 & \bf{60.0} & 75.7 & \bf{82.3} & 64.4 & \bf{73.8} \\
			\hline
		\end{tabular}
	\end{center}
	\caption{%
		MobileNet results for various datasets.
		The columns `orig' refer to the baseline MobileNet models.
		The columns `ours' refer to \DCCKU for MobileNetV1 and \DCCKS for MobileNetV2/V3.
	}
	\label{tab:CIFAR-Mobilenet}
\end{table*}

The results are shown in \Cref{tab:CIFAR-Mobilenet}.
We can state that all \DCCK variants outperform their corresponding MobileNet baselines.
For MobileNetV1, this can be explained by
(i) the inversion of point- and depthwise convolutions and
(ii) the absence of ReLU activations for the pointwise convolutions
(see discussion in \Cref{sec:dcck:mobilenets:dcck-u}).
For \mbox{MobileNetsV2} and V3, the fact that \DCCKS always outperforms the baseline models clearly confirms the advantage of our proposed orthonormal regularization loss.

\textbf{ResNets and WideResNets.}
In addition to the improvements with respect to MobileNets, we can use our \DCCK approach as drop-in substitution for regular convolution layers in standard networks.
In the following, we consider ResNets \cite{he2016deep} and WideResNets \cite{zagoruyko2016wide} as two state-of-the-art models for the CIFAR datasets.
In both cases, we use the \DCCKU variant.
We increase the depth and width factor of each \DCCK model such that its parameter count matches the parameter count of the corresponding baseline model.
We apply the same training protocol and augmentation techniques as described above.

\begin{table*}
	\begin{center}
		\begin{tabular}{|l|c|c|c|c|c|c|c|}
			\hline
			Network & \multicolumn{3}{|c|}{CIFAR10} & \multicolumn{3}{|c|}{CIFAR100} \\
			 &  Parameters & FLOPs & Accuracy & Parameters & FLOPs & Accuracy \\
			\hline
			ResNet-20 \cite{he2016deep} 	& 272.5K & 41.3M & 92.2 & 278.3K & 41.3M & 67.7 \\
			ResNet-110 (\DCCKU) 			& \bf{239.0K} & \bf{41.1M} & \bf{92.9} & \bf{244.8K} & \bf{41.1M} & \bf{70.8} \\
			\hline
			WideResNet-40-3 \cite{zagoruyko2016wide} & 5.0M & 735.0M & 94.9 & 5.0M & 735.0M & 75.5 \\
			WideResNet-40-8 (\DCCKU)	 	& \bf{4.2M} & \bf{671.6M} & \bf{95.2} & \bf{4.3M} & \bf{671.6M} & \bf{77.6} \\
			\hline
			\end{tabular}
		\end{center}
	\caption{%
		ResNets and WideResNets on CIFAR10 and CIFAR100.
		We increase the depth and width factor of each \DCCK model such that its parameter count matches the parameter count of the corresponding baseline model.
	}
	\label{tab:CIFAR-ResNet}
\end{table*}

In \Cref{tab:CIFAR-ResNet} we compare the original networks with their modified \DCCK versions.
For ResNets, we can improve accuracy by up to $3.1$ percentage points for CIFAR100, while having slightly fewer parameters and computational costs.
For WideResNets, we can gain accuracy of up to $2.1$ percentage points for CIFAR100, while having fewer parameters and computational costs.
This clearly shows the effectiveness of our approach as an drop-in substitution of regular convolution layers.

\subsection{ImageNet}
To assess the performance of \DCCK models in large-scale classification scenarios, we conduct experiments on the ImageNet dataset (ILSVRC2012, \cite{russakovsky2015imagenet}).
It contains about 1.3M images for training and 50k images for testing which are drawn from 1000 object categories.

We employ a common training protocol and train for 100 epochs with
an initial learning rate of $0.1$ which is decayed by a factor of $0.1$ at epochs $30$, $60$, and $90$.
We use SGD with momentum $0.9$ and a weight decay of $10^{-4}$.
To allow for a fair comparison and to investigate the effect of our approach, we train own baseline models with exactly the same training setup as used for \DCCK models.
The images are resized such that their short side has a length of $256$ px.
We use the well-established Inception-like scale augmentation \cite{szegedy2015going}, horizontal flips, and color jitter \cite{krizhevsky2012imagenet}.

\textbf{MobileNets.}
As for the CIFAR experiments, we compare MobileNets to their corresponding \DCCK variants.
Again, \DCCKU is used for MobileNetV1, and \mbox{\DCCKS} is used for MobileNetV2/V3.
The subspace compression ratio for \DCCKS is $p = \frac{1}{6}$ just like for the CIFAR experiments.
The weighting coefficient $\alpha$ for the orthonormal regularization loss was set to $0.1$.

\begin{table}
	\begin{center}
		\begin{tabular}{|l|c|c|c|c|}
			\hline
			Network & Original & \DCCK (ours) \\
			\hline
			MobileNetV1 \widthFactor{0.25} 		& 51.8 & \bf{53.2} \\
			MobileNetV1 \widthFactor{0.5}  		& 63.5 & \bf{64.6} \\
			MobileNetV1 \widthFactor{0.75} 		& 68.2 & \bf{69.2} \\
			MobileNetV1 \widthFactor{1.0}  		& 70.8 & \bf{71.5} \\
			\hline
			MobileNetV2 \widthFactor{1.0} 		& 69.7 & \bf{69.8} \\
			\hline
			MobileNetV3-small \widthFactor{1.0} 	& 64.4 & \bf{64.8} \\
			\hline
			MobileNetV3-large \widthFactor{1.0} 	& \bf{71.5} & \bf{71.5} \\
			\hline
		\end{tabular}
	\end{center}
	\caption{%
		MobileNets on ImageNet.
		\DCCKU is used for MobileNetV1, and \DCCKS is used for MobileNetV2/V3.
		Note that \DCCK does not introduce additional parameters.
	}
	\label{tab:ImageNet-Mobilenet}
\end{table}

The results are presented in \Cref{tab:ImageNet-Mobilenet}.
Again, it can be seen that the \DCCK variants of MobileNets outperform their corresponding baseline models.
However, the relative improvements are no longer as large as for the \mbox{CIFAR} experiments.
This effect can be explained by the regularization impact of the dataset itself.
Considering the \mbox{MobileNetV3-large} results, we note that even if the orthonormal regularization loss seems to be no longer effective, it has no negative influence on the training.

\textbf{ResNets.}
As noted before, it is possible to directly substitute regular convolution layers in standard networks by \DCCK variants.
To this end, we analyze the effectiveness of our approach when applied to ResNets on large-scale image databases.
For the baseline models, we use ResNet-10, ResNet-18, and ResNet-26.
The \DCCK variants are ResNet-10, ResNet-18, ResNet-34, ResNet-68, and ResNet-102.
Again, we use the same training protocol and augmentation techniques as described above.

\begin{figure*}
	\center
	\includegraphics[width=0.45\textwidth]{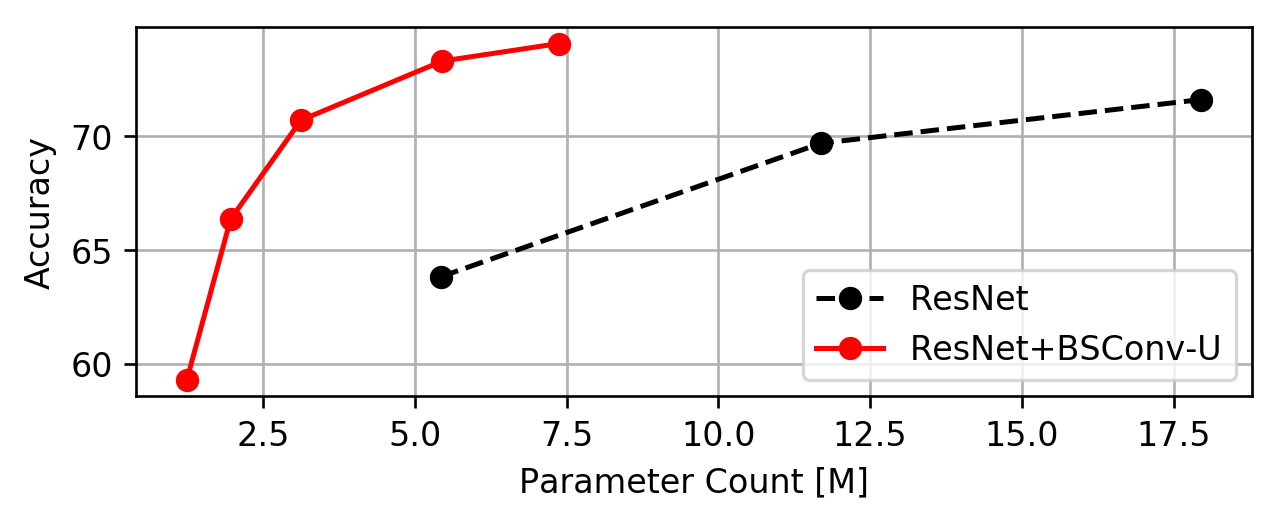}~~~~~~~~
	\includegraphics[width=0.45\textwidth]{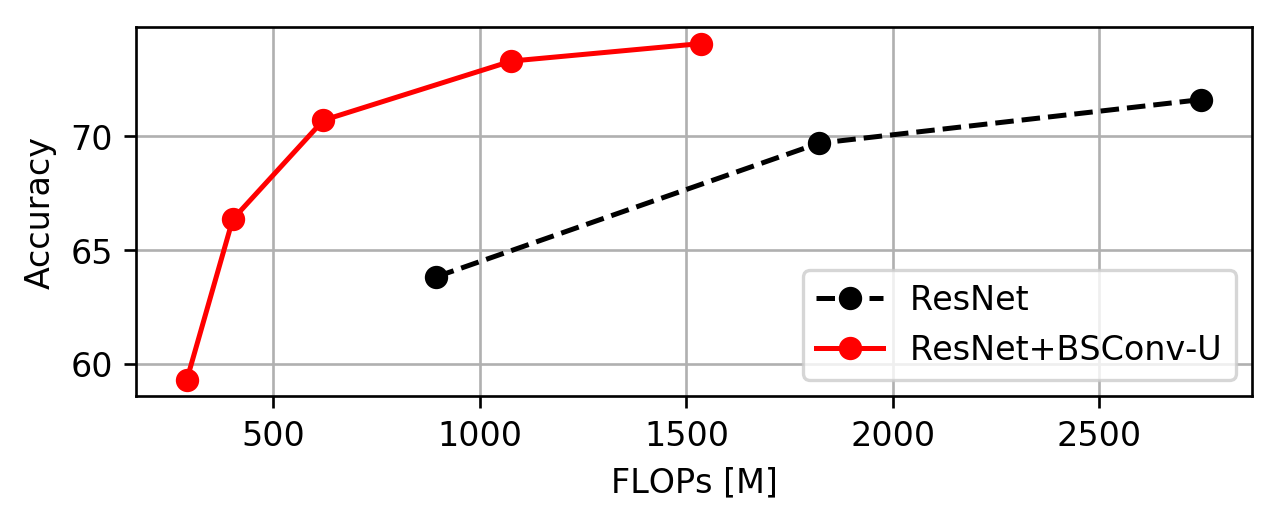}
	\caption{%
		ResNets on ImageNet.
		For the baseline models, we use ResNet-10/18/26.
		The \DCCK variants are ResNet-10/18/34/68/102.
	}
     \label{fig:resnetsImagenetParamsAndFlops}
\end{figure*}

The results are shown in \Cref{fig:resnetsImagenetParamsAndFlops}, split by parameter count and computational complexity.
It can be seen that the \DCCKU variants of ResNets significantly outperform the baseline models.
ResNet-10 and \mbox{ResNet-68+\DCCKU}, for instance, have similar parameter counts, while using \DCCK leads to an accuracy gain of $9.5$ percentage points.
Another interesting example is ResNet-18 vs. \mbox{ResNet-34+\DCCKU}: both have a comparable accuracy, while the \DCCK model has only about one fifth of the baseline model parameter count.

\subsection{Fine-grained Recognition}
Apart from large-scale object recognition, we are interested in the task of fine-grained classification, as those datasets usually have no inherent regularization.
The following experiments are conducted on three well-established benchmark datasets for fine-grained recognition, namely Stanford Dogs \cite{Khosla2011dogs}, Stanford Cars \cite{KrauseStarkDengFei-Fei_3DRR2013}, and Oxford 102 Flowers \cite{nilsback2008automated}.
We train all models from scratch, since parts of these datasets are a subset of ImageNet.
In contrast to the ImageNet training protocol, we do not use aggressive data augmentation, since we observed that it severely affects model performance.
We only augment data via random crops, horizontal flips, and random gamma transform.

We use the same training protocol for all three datasets.
In particular, we use SGD with momentum set to $0.9$ and a weight decay of $10^{-4}$.
The initial learning rate is set to $0.1$ and linearly decayed at every epoch such that it approaches zero after a total of $100$ epochs.

\textbf{MobileNets.}
We use the same model setup as for the \mbox{CIFAR} and ImageNet experiments discussed above.
The results are shown in \Cref{tab:CIFAR-Mobilenet}.
Again, all \DCCK models substantially outperform their baseline counterparts.
In contrast to the CIFAR results, the margin is even larger.
Therefore, the interpretation of the CIFAR results applies here as well.

\textbf{Other Architectures.}
We further evaluate the effect of our approach for a variety of state-of-the-art models.
We replace regular convolution layers in standard networks such as VGG \cite{simonyan2014very} and DenseNet \cite{huang2017densely}.

In \Cref{tab:dogsOthers} we can see that all models greatly benefit from the application of \DCCK.
Accuracy for \DCCKU can be improved by at least 2 percentage points, while having up to $8.5 \times$ less parameters and a substantial reduction of computational complexity.
Most of the recently proposed model architectures utilize residual linear bottlenecks \cite{sandler2018mobilenetv2}, which can also be easily equipped with our \DCCKS approach in the same way as for MobileNetV2/V3 (see \Cref{sec:dcck:mobilenets:dcck-s}).
As can be seen in \Cref{tab:dogsOthers}, our subspace model clearly outperforms the original EfficientNet-B0 \cite{tan2019efficientnet} by $6.5$ percentage points and MnasNet \cite{tan2019mnasnet} by $5$ percentage points with the same number of parameters and computational complexity.
This shows the effectiveness of our proposed orthonormal regularization of the \DCCKS subspace transform.

\begin{table}
	\begin{center}
		\begin{tabular}{|l|c|}
			\hline
			Network & Accuracy \\
			\hline
			VGG-16 (BN) \cite{simonyan2014very}	& 60.5 \\
			VGG-16 (BN) (\DCCKU) 				& \bf{62.4} \\
			\hline
			DenseNet-121 \cite{huang2017densely}	& 56.9 \\
			DenseNet-121 (\DCCKU) 				& \bf{59.4} \\
			\hline
			Xception* \cite{chollet2017xception}	& 59.6 \\
			Xception (\DCCKU) 					& \bf{64.3} \\
			\hline
			EfficientNet-B0 \cite{tan2019efficientnet}	& 54.7 \\
			EfficientNet-B0 (\DCCKS)				& \bf{61.2} \\
			\hline
			MnasNet \cite{tan2019mnasnet}			& 54.8 \\
			MnasNet (\DCCKS)						& \bf{59.8} \\
			\hline
			\end{tabular}
		\end{center}
	\caption{%
		Results of various architectures and their \DCCK counterparts for the Stanford Dogs dataset.
		\DCCKU CNNs have fewer parameters and a smaller computational complexity compared to their baseline models.
		\DCCKS CNNs have the same parameter count and computational complexity as their counterparts.
		* Commonly used implementation based on DSCs.
	}
	\label{tab:dogsOthers}
\end{table}

\textbf{Influence of the Orthonormal Regularization.}
To evaluate the influence of the proposed orthonormal regularization loss for \DCCKS models, we conduct an ablation study using MobileNetV3-large.
In particular, several identical models are trained on the Stanford Dogs dataset using weighting coefficients $\alpha$ in the range of $10^{-5}, \ldots, 10^{0}$.

\begin{figure}
	\center
	\includegraphics[width=1.0\columnwidth]{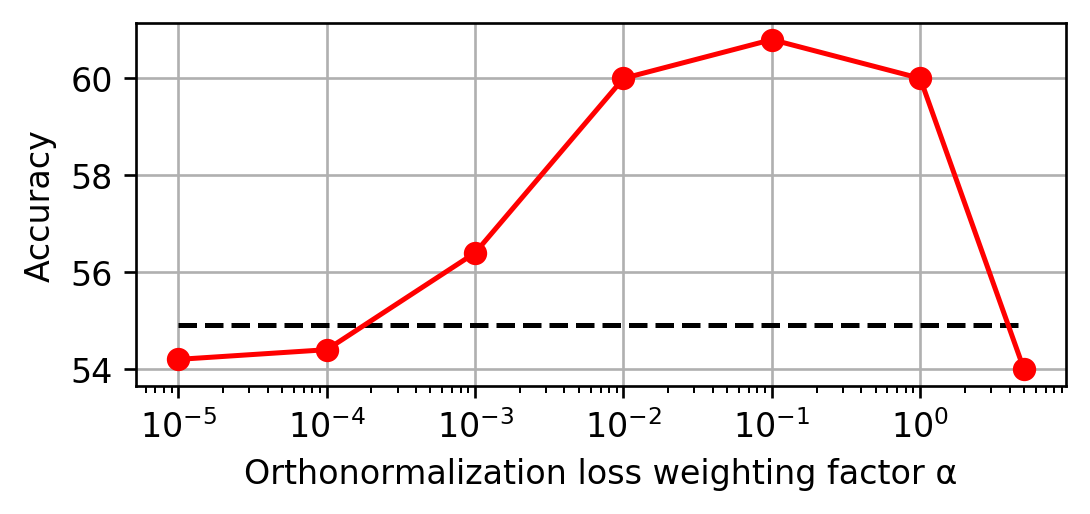}
	\caption{%
		Influence of the orthonormal regularization loss on the accuracy for the \DCCKS variant of MobileNetV3-large (red solid line) on Stanford Dogs.
		The baseline MobileNetV3-large model without \DCCKS is indicated by the black dashed line.
	}
     \label{fig:dcckV2Alpha}
\end{figure}

As can be seen in \Cref{fig:dcckV2Alpha}, by regularizing the subspace components to be orthonormal, model performance can be substantially improved by over $5$ percentage points.
An optimum is reached for a weighting coefficient of $\alpha=0.1$.
For smaller values, the influence of the regularization decreases, until it is no longer effective and converges towards the baseline performance.
Larger values, however, decrease model performance since the optimization is mainly driven by rapidly reaching a solution with an orthonormal basis independently of creating a beneficial joint representation.

%%%%%%%%%%%%%%%%%%%%%%%%%%%%%%%%%%%%%%%%
% conclusions
%%%%%%%%%%%%%%%%%%%%%%%%%%%%%%%%%%%%%%%%

\section{Conclusions}
We introduced blueprint separable convolutions (\DCCK) as highly efficient building blocks for CNNs.
Our formulation provided an interpretation and justification for depthwise separable convolutions.
By using \DCCK, we clearly and consistently improved established models such as MobileNets, MnasNets, EfficientNets, and ResNets.
Code and trained models are available under \url{https://github.com/zeiss-microscopy/BSConv}. 

%%%%%%%%%%%%%%%%%%%%%%%%%%%%%%%%%%%%%%%%
% bibliography
%%%%%%%%%%%%%%%%%%%%%%%%%%%%%%%%%%%%%%%%

{\small
\bibliographystyle{ieee_fullname}
\bibliography{main}
}

\end{document}